\documentclass[letterpaper]{article}
\usepackage{proceed2e}
\usepackage[margin=1in]{geometry}

\usepackage{microtype}
\usepackage{graphicx}
\usepackage{subfigure}
\usepackage{booktabs} % for professional tables
\usepackage{amsmath}
\usepackage{amsfonts}
\usepackage[utf8]{inputenc}
\usepackage[english]{babel}
\usepackage{amsthm}
\usepackage{xcolor}
\usepackage{natbib}

\newtheorem{MyProperty}{Property}

\def\url#1{\expandafter\string\csname #1\endcsname}
\usepackage{times}

\title{A Unified Probabilistic Model for Learning Latent Factors and \\ Their Connectivities from High-Dimensional Data}

\author{}
\author{ {\bf Ricardo Pio Monti}${^1}$ \and \bf{Aapo Hyv\"{a}rinen}$^{1,2}$ \\
$^1$Gatsby Computational Neuroscience Unit, University College London, UK\\
$^2$Department of Computer Science and HIIT, University of Helsinki, Finland
}

\begin{document}

\maketitle

\begin{abstract}
%	Inferring latent factors from high-dimensional data is fundamental in many applied domains. 
%	
%	
%	Inferring connectivity structure over latent variables is necessary in many applications. 
%	However, two 
%	
%	In many applications it is necessary to infer connectivity structure over latent variables

%	Connectivity estimation is a fundamental problem in many fields of science. 
%	However, in the context of  high-dimensional data it may be
%	neither feasible nor useful to model the connectivities between the original variables. Grouping %the 
%	variables into clusters or communities is a useful preprocessing step, but it is not clear how to do it optimally in view of connectivity estimation. Another practical problem is that we may have data from different classes (e.g.\ multiple subjects or conditions in a biomedical experiment), and we need to incorporate useful constraints about the similarities between the classes. 
%	
%	Here, 
	Connectivity estimation is challenging in the context of high-dimensional data. A useful preprocessing 
	step is to group variables into clusters, % or communities, 
	however, 
%	it is not always clear how to perform such a grouping.
	 it is not always clear how to do so %it may not be clear how to do so 
%	it is not clear how to cluster variables %identify the optimal clusters 
	from the perspective of 
%	it is not clear how to do so
	connectivity estimation.  
	Another practical challenge is that we may have data from multiple 
	related classes (e.g., multiple subjects or conditions) and wish to incorporate 
	constraints on the similarities across classes.
	We propose a  probabilistic model which simultaneously performs both a grouping of variables (i.e., detecting community structure) and estimation of connectivities between the groups which correspond to latent variables. 
	The model is essentially a  factor analysis model where the factors are allowed to have arbitrary correlations, while the factor loading matrix is constrained to express a community structure. The model can be applied on multiple classes
 so that the connectivities can be different between the classes, while the community structure is the same for all classes. We propose an efficient estimation algorithm based on score matching, and prove the identifiability of the model.
% , i.e.,
%the uniqueness of the solution. 
%{\color{black} 
	Finally, 
	we present an extension to directed (causal) connectivities over latent variables.
%	which first clusters variables according to the aforementioned model. Causal structure can 
%	then be inferred using existing methods.
%}
	%	 	which 
	%	 	
	%	 	We further present an extension to causal latent variable models where
	%	 	the goal is to learn causal structure over latent variables. 
	%		the distribution over latent variables is assumed to follow a directed 	 
Simulations and experiments on fMRI %brain imaging 
data validate the practical utility of the method.
\end{abstract}

%\section{Introduction}
\section{INTRODUCTION}
Estimating the connectivity structure between observed variables is a
fundamental problem in statistics and machine learning.
Probabilistic methods are often based on estimation of the covariance
matrix or it inverse. A number of estimators have been proposed for
both \citep{Dempster1972, Ledoit}. 
On a more
general level, such
{\color{black} undirected }
connectivity estimation is a special case of modelling
the covariance matrix, which is one of the goals of classical
dimensionality reduction methods such as factor analysis and
principal components analysis (PCA).
{\color{black}
In contrast, directed connectivity estimation 
studies the causal dependence structure across variables \citep{Pearl2009}.
In this work we present methods to perform both 
directed and undirected connectivity estimation of latent variables in 
the context of high-dimensional data.
%
%We focus primarily on undirected connectivity, but 
%
%the methods developed in this work are also applied
%also present 
%experiments demonstrating the utility of the proposed method in the context of 
%causal methods \citep{Shimizu2006, Silva2006}.	
}
%We constrain ourselves here to undirected connectivity,
%in contrast to causal methods \cite{Shimizu2006}. 

An important problem in practice is that many connectivity estimation methods assume we observe, and know how to choose, the variables between which the connectivity is to be estimated. However, in practice we often have very high-dimensional data, and it may not be useful or feasible to estimate the connectivities between all of them. It is important to somehow reduce the number of variables so that the connectivity estimation is feasible, and furthermore, such reduction can greatly facilitate interpretation of the results. 
%In network science, finding community structure is a prominent topic, while, for example, in brain imaging, some kind of parcellation is ubiquitous. 
{\color{black}
%	It is often desirable to combine dimension reduction and clustering
It is often useful to perform the dimension reduction so that it can be interpreted as clustering,
% of the observed variables, 
as in non-negative PCA \citep{Sigg2008}. 
%
%\newpage 	
%From the perspective of undirected connectivities, 
%a closely related method is non-negative PCA \citep{Sigg2008, Zass2007}. %, but
%However, 
A 
relevant challenge is how such reduction in the number of variables should be 
combined with connectivity estimation. In the past, approaches based on
stochastic block models \citep{Airoldi2008, Marlin2009}
or clustered factor analysis \citep{Buesing2014}
have been employed.
However, such methods
do not explicitly model the connectivity over latent variables and 
cannot easily be extended to accommodate multiple classes of related datasets, both of which are
of interest in this work. 
Alternative methods
recover correlation structure over latent variables but do not focus on dimensionality
reduction \citep{Sasaki17NC}. 
%primary focus of this work. % as we discuss below. 
% are computationally
%expensive and cannot easily be extended to accommodate multiple classes. 
%An open problem is, however, how  such reduction in the number of variables should be optimally combined with connectivity estimation.
Conversely, in the context of directed connectivity, the 
causal clustering of observed variables has been studied by
%Whereas in the context of directed connectivities this has been studied by 
\citet{Silva2006}, \citet{Shimizu2008} and \citet{Kummerfeld2016}.
%Whereas in the context of directed connectivities, related methods include 
%
%
%An important difference is that we enforce both non-negativity and orthonormality, whereas the latter constraint is typically 
%relaxed. % in the context of non-negative PCA as it is too restrictive 
}

A further challenge is estimating multiple
related connectivity matrices, assuming the data is divided into a
number of classes, such as subjects in a biomedical setting.  While the estimation of multiple related Gaussian
graphical models, parameterized by the inverse covariance, has been
extensively studied
\citep{Varoquaux2010, Danaher2014, Monti2017}, we want to combine such multiple connectivity
estimation with the variable reduction scheme described above
{\color{black}
as well as extend such methods to the domain of directed connectivities.
}

A practical application that motivates our theoretical
developments is functional MRI (fMRI) data, where estimation of ``functional
connectivity'' is widely practiced. Such analysis is a cornerstone of modern neuroscientific
research, having provided
%which has provided 
fundamental insights into the 
structure and architecture of the human connectome. However, the existing methods are often not very rigorous and would benefit from a proper probabilistic formulation.
Traditionally, functional connectivity networks have been modeled as covariance graphs, where the 
nodes in the network correspond to spatially remote brain regions and 
edges encode the marginal dependence structure (often simply the covariance).
In fMRI, we very clearly see the importance of the theoretical points raised above, in the form of the following challenges:
\begin{itemize}
	\item {{\color{black}Inter-subject consistency}}: Data is often collected across a cohort of subjects. 
	A hallmark of brain networks is their inter-subject consistency; 
	observed patterns in connectivity have been shown to demonstrate reproducible properties 
	across subjects \citep{Damoiseaux2006}. This suggests 
	significant benefits can be obtained by sharing information across subjects in 
	a judicious manner. In fact, some of the most recent developments are based on collecting hundreds or even thousands of subjects' data in a single data base \citep{DiMartino2014}.
	%	Many current methods either ignore the reproducible property of
	%	brain networks (and estimate a covariance graph for each subject independently) or 
	%	na{i}vely aggregate data across all subjects, thereby precluding the study of differences across subjects.
	\item {Modularity}: 
	Current methods do not actively incorporate  
	domain knowledge relating to brain networks, a prime example of which is their 
	modular structure 
	which suggests that variables can be aggregated into non-overlapping modules or 
	sub-networks
	\citep{Sporns2016}. 
	We note 
	this property is not unique to brain networks, but also present in 
	many real-world networks
%	Moreover, real-world networks frequently display modular structure where nodes can easily be divided into
%	modules 
	\citep{Newman2006}. 
\end{itemize}

While motivated by fMRI,
we note that these two  properties are relevant to wide range of applications such as cyber-security, gene expression data and econometrics.

In this work, we propose a probabilistic latent variable model which
is able to directly address the aforementioned issues.  The proposed
model consists of a low dimensional set of latent variables in a
factor analytic model.  The associated factor loading matrix is shared
across classes and constrained to be non-negative and orthonormal,
thereby encoding module/community membership along its columns.  Thus,
the factors are interpreted as the activities in modules or
communities.

Importantly, and in contrast to almost all related models,  these latent variables or factors have full (i.e., non-diagonal) covariance structure which we term \textit{latent connectivities}, giving the connectivity structure of the  
non-overlapping modules. The connectivity structure can be different between classes; however, the model can equally well be applied on data from a single class. Thus, we model both the grouping of variables, and the connectivity between the groups in a single probabilistic model, which can be seen as a variant of factor analysis.  We argue that such a formulation leads to important benefits from the viewpoint of interpretation 
{\color{black}
and identifiability 
whilst remaining plausible from an
application perspective.
}
%as well as hypothesis testing (since
%we are able to interpret changes in network structure on a
%module-by-module basis), whilst remaining plausible from an
%application perspective.  

In contrast to classical Gaussian factor analysis, we are able to prove the uniqueness of the solution: the factors and loadings are identifiable like in (non-Gaussian) independent component analysis  \citep{Comon1994}, largely based on non-negativity inherent in the module structure  \citep{Paatero1994,Seung1999,Donoho2004}.
We further propose an efficient parameter estimation algorithm based on score
matching. 
{\color{black}
Finally, we demonstrate that the proposed model can be
extended to modelling directed connectivities between the latent variables. In this context, the 
factor loading matrix can be seen as a ``pure'' measurement model of a Bayesian network \citep{Silva2006}
of causal relationships between high-dimensional observations 
and their latent variables. 
%Given a correct measurement model, there is a wide 
%literature available to infer the causal structure across latent variables \citep{Silva2006}. 
% estimate the 
%\textit{measurement model} between 
%	
%Finally, we present a simple extension of the proposed method to 
%learn causal relations across latent variables. 
}
%We  demonstrate the capabilities of the model using simulations 
%and experiments on resting-state fMRI data.

{\color{black}
The remainder of this manuscript is organized as follows;
in Section \ref{sec--LatentConnModel} we present the 
proposed model in the context of undirected latent variables. 
Section \ref{sec--Identifiable} provides an identifiability analysis for the proposed method.
%while 
An efficient estimation algorithm based on score matching is 
presented in Section \ref{Sec--SMalgo}. 
The proposed method is extended to recover causal structure over latent variables in 
Section \ref{sec--ExtensionDAG}. %Simulations and 
Experimental results %on resting-state fMRI data
 are presented in Section \ref{sec--Experiments}. 
}

%\section{PRELIMINARIES}

%\newpage 

\section{LATENT CONNECTIVITIES MODEL}
% \section{Latent connectivities model}
\label{sec--LatentConnModel}

We propose a latent variable model to 
accurately find modules (communities, clusters) and model their connectivities, possibly across multiple related 
classes (conditions, subjects).
%Crucially, we assume that observations within each class are exchangeable 
%but 
We assume we have access to multivariate data over $N$ distinct 
classes, but all our results allow for  the simple case  $N=1$ as well.
For a given class $i$,
we write $X^{(i)} \in \mathbb{R}^{p}$ to denote the $p$-dimensional observed random vector. 
The $i$th class is associated with 
a $k$-dimensional latent vector, $Z^{(i)}$,
which is related to observations, $X^{(i)}$, via a loading matrix $W \in \mathbb{R}^{p \times k}$. 
We note that the loading matrix is shared across all classes and will serve to encode 
 module memberships across classes.

We start by a model of undirected connectivities in this section.
Here, we assume that the data for each class follows a stationary 
multivariate Gaussian distribution with zero mean and  covariance $\Sigma^{(i)} \in \mathbb{R}^{p \times p}$.
Both observations and latent variables are taken to follow multivariate Gaussian distributions, such that:
%We further assume that both $Z^{(s)}, X^{(s)}$ are Gaussian, resulting in a 
%Gaussian latent variable model: % this corresponds to a 
\begin{align}
\label{FA_eq1}
Z^{(i)} &\sim \mathcal{N} \left (0, G^{(i)}  \right ) \\
X^{(i)} | Z^{(i)} = z^{(i)} &\sim \mathcal{N}\left (W z^{(i)}, v^{(i)} I  \right ). 
\label{FA_eq2}
\end{align}

\begin{figure}[t!]
	\centering
	\includegraphics[width=.5\textwidth]{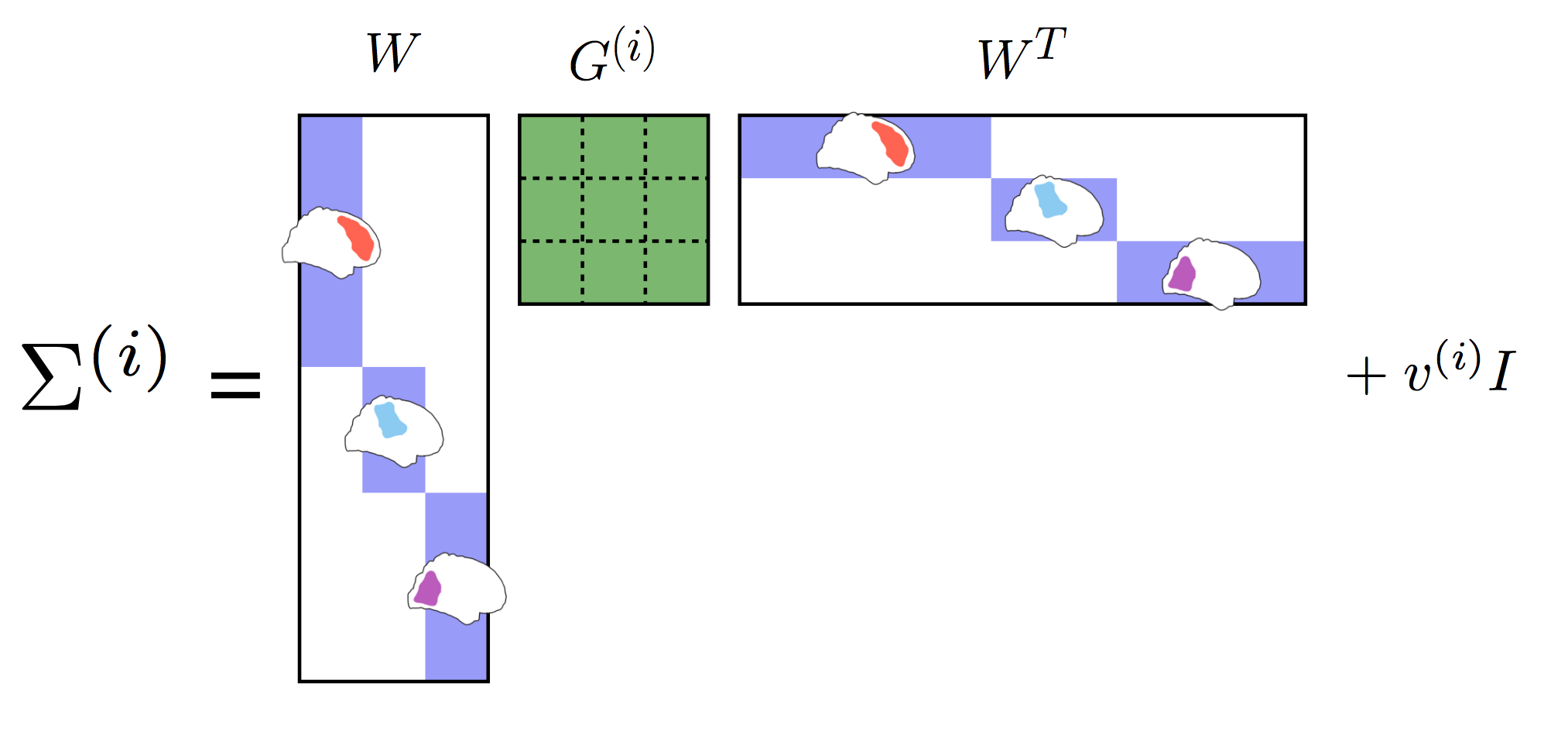}%{MotivatingFig.pdf}
	\caption{Visualization describing  the various components of the proposed covariance model. 
		The factor loading matrix, $W$, is shared across all subjects and 
		serves to denote membership into non-overlapping brain modules. 
		The \textit{latent connectivity} across modules, parameterized by $G^{(i)}$, is allowed to 
		vary across subjects. 
%		This model therefore assumes that while modules  are consistent  and reproducible across
%		classes, the connectivity across modules may vary significantly. 
		%		
		%		The non-zero entries of $W$ denote membership to non-overlapping brain modules which are
		%		consistent across all subjects. The connectivity across modules is parameterized by $G^{(s)}$
		%		and is allowed to vary across subjects. 
		%		We mode the functional connectivity for subject, $\Sigma^{(s)}$, as the 
		%	product of shared modules, $W$, and a subject-specific inter-module connectivity matrix $G^{(s)}$.
		%We model FC as being decomposed into $k$ modules, denoted by the non-zero entries of $W$.  
	}
	\label{fig:Motivating}
\end{figure}

If  $G^{(i)}$ were diagonal,
equations (\ref{FA_eq1}) and (\ref{FA_eq2}) would correspond to the traditional factor analysis or probabilistic
PCA models \citep{Tipping1999}. 
Our model is able to capture low-rank covariance structure via the loading matrix, $W$, as follows:
\begin{equation}
\label{CovModelEq}
%cov \left ( X^{(s)} \right ) 
\Sigma^{(i)} = W G^{(i)} W^T + v^{(i)} I.
\end{equation}
From equation (\ref{CovModelEq}) it follows that the loading matrix $W$ serves to encode reproducible 
covariance structure which is present across all classes.
In this work, we extend the traditional factor analysis model as follows:
\begin{itemize}
	\item The loading matrix, $W$, is constrained to be non-negative and orthonormal. 
	This leads to a loading matrix 
	with at most one non-zero entry per row.  We may interpret the columns of $W$ as 
	encoding membership to $k$ non-overlapping modules or sub-networks. 
	\item We introduce latent variables with a non-diagonal covariance structure, which we term
	\textit{latent connectivities}. 
	While the columns of the loading matrix encode module membership, the 
	non-trivial covariance structure over latent variables may be interpreted as 
	modeling marginal dependencies (i.e., connectivity) across distinct modules or sub-networks. 
	Such an interpretation is 
	very natural in many applied settings. 
\end{itemize}
We note that the introduction of marginally dependent latent variables is 
not possible in the context of 
traditional factor analysis, since the effects of factor connectivity and factor loadings cannot be distinguished. In fact, an ordinary factor analysis model is non-indentifiable even with uncorrelated factors.
However, 
in combination with the aforementioned constraints on the loading matrix, it is possible to identify the latent connectivities in our model (see next section).
We thus argue that our model 
is able to capture the modular nature of many real-world datasets and, due to its identifiability, 
yields easily interpretable results. 
Figure \ref{fig:Motivating} provides an overview of the proposed model in the context 
of estimating brain connectivity networks. 
%it is desirable in our case as we seek to  estimate multiple covariance matrices. 

%\section{Identifiability analysis}
\section{IDENTIFIABILITY ANALYSIS}
\label{sec--Identifiable}

We note that 
without the introduction of constraints on $W$, the 
covariance model proposed in equation (\ref{CovModelEq}) 
is not unique. For example, 
%in the case of a single subject ($S=1$) 
it would be possible to re-parameterize $W$ 
such that $G^{(i)}$ is diagonal matrix by using an eigen-value decomposition of  $G^{(i)}$. 
%Such an approach is less feasible in the context of many subjects whose latent 
%variables have distinct covariances. 
%Furthermore, as the two properties below 
%provide some local identifiability guarantees for the proposed model. 
%However, the introduction of non-negativity and orthonormality constraints on 
%the loading matrix ensure the problem is identifiable. 
However, the following properties 
demonstrate that non-negativity and orthonormality constraints 
are sufficient to 
ensure the solution is identifiable.

\begin{MyProperty}
	\label{PropP1}
	{Assume non-negative, orthonormal $W$. Then at most one entry 
		per row of $W$ can be non-zero.}
\end{MyProperty}
%\MyProperty{Assume non-negative, orthonormal $W$. Then at most one entry 
%per row of $W$ can be non-zero.}
\begin{proof}
	Directly from the constraints on $W$, we can express the $(i,j)$ entry of $W^TW$ as:
	%\begin{align*}
	$$
	\left (W^T W \right )_{ij} = \sum_{r=1}^k \left ( W^T\right)_{ir} W_{rj}  = \sum_{r=1}^k W_{ri} W_{rj} 
	=\delta_{ij} 
	$$
	% \end{align*}
	which, combined with non-negativity, implies that the $i$ and $j$ columns of $W$ can have no overlapping support for $i \neq j$. 
\end{proof}

\begin{MyProperty}
	\label{PropP2}
	Assume non-negative, orthonormal $W$.  
	Then any matrix $V  \in \mathbb{R}^{k \times k}$ 
	for which 
	$\tilde W = W V$
	is non-negative and orthonormal must be the identity matrix or some permutation of the 
	identity. 
	%	Wtilde=WV is non-negative and orthonormal  must be the identity matrix o
	%	Then there exists $V \in \mathbb{R}^{k \times k}$ 
	%	such that $\tilde W = W V$ and $\tilde W$ is also non-negative and orthonormal 
	%	if and only if $V$ is the identity matrix or some permutation of it.   
\end{MyProperty}
\begin{proof}
	By Property 1 we have that both $W$ and $\tilde W$ have at most one non-zero entry per 
	row. Since $\tilde W^T \tilde W = I$ we have that $V^T V=I$. 
	%	
	%	We consider the $i$th row of $W$ and $\tilde W$. 
	Define $c_i$ and $\tilde c_i$ to be  the 
	index of the non-zero entry along the $i$th row for 
	$W$ and $\tilde W$ respectively. 
	By construction:
	$$ \tilde W_{i \tilde c_i} = \left (W V\right)_{i \tilde c_i}  =\sum_{r=1}^k  W_{ir} V_{r \tilde c_i}  = W_{i c_i} V_{c_i \tilde c_i}  >  0$$
	where the final equality follows from the fact that
	$W_{i c_i}$ is the only non-zero entry along the $i$th row of $W$. 
	Since $W_{i c_i} > 0$, this implies that $V_{c_i \tilde c_i} >0$. 
	Furthermore, for $j \neq \tilde c_i$ we have
	$$ \tilde W_{i j} = \left (W V\right)_{i j}  =\sum_{r=1}^k  W_{ir} V_{r j}  = W_{i c_i} V_{c_i j}  =  0$$
	Since $W_{i c_i} > 0$, we must have that $V_{c_i j} =0$ whenever $j \neq \tilde c_i$. 
	When combined with the fact that $V^T V = I$, it follows that $V$ must either be the identity matrix
	of a permutation it.
\end{proof}

Property \ref{PropP2} 
indicates that the 
matrix $W$ is uniquely defined in our model, and there is nothing like an undetermined factor rotation in conventional Gaussian factor analysis. 
By similar logic, Property \ref{PropP2} also
%property further 
implies the uniqueness of $G^{(i)}$. 

%\section{Estimation by  score matching}
\section{ESTIMATION BY SCORE MATCHING}
\label{Sec--SMalgo}

The parameters associated with the proposed model consist of
the loading matrix, $W$, the latent variable covariances, $\{ G^{(i)} \}$, and the 
observation noise, $\{v^{(i)} \}$. 
%observation noise $\sigma^2$ which is assumed shared across subjects for simplicity. 
One potential strategy is to estimate latent variables 
in an expectation-maximization framework. However, 
due to relative simplicity of the proposed covariance model 
%in this work 
we propose to directly
marginalize out latent variables. 

Parameters may also be  estimated 
via maximum likelihood estimation, however, this 
{\color{black}
results in an iterative algorithm where the computational cost of each 
parameter update is
}
%results in an algorithm with a computational cost of  
$\mathcal{O}(p^3)$ (a derivation of which is provided in the Supplementary materials).
Instead, we propose to estimate parameters by 
score matching \citep{Hyvarinen2006}, leading to an algorithm with a computational cost of 
$\mathcal{O}(p^2 k)$ {\color{black} per iteration}. This is a significant reduction
as we will typically expect $k \ll p$.
{\color{black} 
While score matching is typically used in the context of unnormalized statistical models,
it may often result in optimization-related benefits for normalized models as well \citep{Hyvarinen2007, Lin2016}.}
%We note that 
%while score matching was motivated 
%by the estimation of unnormalized statistical models, its 
%has also been employed in order to obtain optimization related benefits, as is the case here.
%For example, efficient score matching algorithms for 
%exponential family models were first derived by \citet{Hyvarinen2007}
%and have been recently employed for regularized estimation by \citet{Lin2016}. 

In the context of multivariate Gaussian data, the score matching objective function is 
defined as \citep{Hyvarinen2006}:
\begin{equation}
\label{SMobj}
\footnotesize
%J\left (W, \{G^{(i)}\} , \{v^{(i)}\} \right )
J = \sum_{i=1}^N - \mbox{tr} \left ( \Omega^{(i)} \right ) + \frac{1}{2} \mbox{tr} \left ( \Omega^{(i)}  \Omega^{(i)}  K^{(i)} \right ),
\end{equation}
where $K^{(i)}$ is the sample covariance matrix for class $i$ and 
$ \Omega^{(i)} $ is the inverse covariance, which
may be computed 
%is defined 
by the Sherman-Woodbury identity as:
\begin{align*}
\footnotesize
\Omega^{(i)}  &
%= {\Sigma^{(i)}}^{-1} 
= {v^{(i)}}^{-1} \left ( I -  W G^{(i)} (G^{(i)} + v^{(i)} I)^{-1} W^T \right ). 
\end{align*}
%where we define $A^{(s)} = G^{(s)} (G^{(s)} + \sigma^2 I)^{-1} $. 
%In the remainder we write $A^{(s)}

Directly optimizing
the score matching objective (equation (\ref{SMobj})) under  non-negativity and orthonormality constraints 
on the loading matrix is challenging. One potential strategy is to 
employ projected gradient descent as suggested by \citet{Hirayama2016}.
However, projecting onto the non-negative Stiefel manifold is undesirable as it requires 
$W$ to have at most one non-zero entry per row at each step of the optimization algorithm (see Property \ref{PropP1} above). Such 
an approach is therefore highly dependent to the random initialization of the loading matrix. 

In this work 
we seek to minimize equation (\ref{SMobj})
in  a constrained optimization framework. This allows for the orthonormality constraints to
be enforced via an augmented Lagrangian penalty \citep{Bertsekas2014}, while the 
non-negativity is enforced at each iteration by projecting onto the non-negative orthant. 

The objective function associated with the augmented Lagrangian is defined as:
\begin{align*}
\label{AugLagrangeObj}
%\min_{W , G^{(s)}, v^{(s) }} \Large \{    J(W, \{G^{(s)} \}) &+ \frac{\rho}{2} || W^T W - I_k ||_2^2 \\
%& + \mbox{tr} (\Lambda^T (W^T W - I_k)) \},
\tilde J =    J + \frac{\rho}{2} || W^T W - I_k ||_2^2 + \mbox{tr} (\Lambda^T (W^T W - I_k)),
\end{align*}
where $\Lambda \in \mathbb{R}^{k \times k}$ are Lagrange multipliers 
enforcing the orthonormality constraints,
$\rho$ is a positive scalar parameter parameterizing the 
augmented penalty term  
and $J$ %$J(W, \{G^{(s)} \})$ 
is the original score matching objective.
We may then proceed to iteratively optimize 
each of the parameters using gradient descent. In particular, the gradient of the score matching objective 
with respect
to the loading matrix can be computed as:
% to each of the parameters can be readily computed as:
\begin{align*}
\frac{\partial J}{\partial W} &= - \sum_{i=1}^N K^{(i)} W A^{(i)} \left  ( I - \frac{1}{2} A^{(i)} \right ),
%\frac{\partial J}{\partial G^{(i)}} &= 	v^{(i)} I - W^T  K^{(i)} W \\&+ W^T K^{(i)}W G^{(i)} ( G^{(i)} + v^{(i)}I )^{-1} \\
\end{align*}
where we define $A^{(i)} = G^{(i)} ( G^{(i)} + v^{(i)}  I )^{-1}$. Similarly, in the case of the latent
connectivities, $G^{(i)}$, and observation noise, $v^{(i)}$, we have
\begin{align*}
\frac{\partial J}{\partial G^{(i)}} = {v^{(i)}}^{-2} &\left  [  {v^{(i)}} I -   W^T K^{(i)} W \left ( I - A^{(i)}\right )  \right  ] \times  \\
& \left [   ( G^{(i)} + v^{(i)} I  )^{-1} \left ( I - A^{(i)}\right ) \right ].\\
\frac{\partial J}{\partial v^{(i)}}  = \sum_{i=1}^N - &{v^{(i)}}^{-3} \mbox{tr} \left ( K^{(i)} - {v^{(i)}} I\right ) \\
&+  {v^{(i)}}^{-3} \mbox{tr} \left ( W^T K^{(i)} W H^{(i)}_1  \right ) + H^{(i)}_2
\end{align*}
where $H^{(i)}_1 \in \mathbb{R}^{k\times k}$ and $H^{(i)}_2 \in \mathbb{R}$ are  defined, together with the relevant derivations,
in the Supplementary material.

Estimation proceeds by iteratively updating each of the 
parameters in a gradient descent framework.
In the context of the loading matrix %, $W$, 
%and 
%observation noise, $\{v^{(i)}\}$, 
%the step-size
the step-size, $\eta$,
is selected via the Armijo rule and we 
project onto the non-negative orthant at each iteration, resulting in an update of the form:
{\small 
\begin{equation*}
W \leftarrow \mathcal{P}_+ \left (  W- \eta   \left (\frac{\partial J}{\partial W } + \rho(W W^T W - W) + W \Lambda \right  ) \right ) 
\end{equation*} 
}where $\mathcal{P}_+ (x) = \mbox{max}(0,x)$ is the projection onto the non-negative orthant. 
%\normalsize 
%$$ W \leftarrow \mathcal{P}_+ \left (  W- \eta   \left (\frac{\partial J}{\partial W } + \rho(W W^T W - W) + W \Lambda \right  ) \right ) 
%$$
Moreover, in the case of latent variable connectivities we have: 
{\scriptsize %\footnotesize
\begin{align*}
\frac{\partial J}{\partial G^{(i)}} =0 \iff G^{(i)} ( G^{(i)} + v^{(i)}I )^{-1} = I - v^{(i)} (W^T K^{(i)} W) 
\end{align*}}which after some manipulation 
 yields a closed form update for the latent connectivity structure as:
\begin{equation}
\label{Gupdate}
G^{(i)} \leftarrow W^T K^{(i)}  W - 	v^{(i)} I 
\end{equation}
By writing $  W^T K^{(i)}  W = (X^{(i)} W)^T (X^{(i)} W)  $
we note that this update has an intuitive interpretation 
as the covariance across  estimated modules. 
%Finally, 
The Lagrange multipliers are updated as \citep{Bertsekas2014}:
$$	\Lambda \leftarrow \Lambda + \rho (W^T W - I)$$

%It is important to note that since the proposed method 
It is important to note that the proposed method only enforces orthonormality 
on the loading matrix in the limit of convergence. However,
the updates provided above are premised on the 
assumption that $W$ is orthonormal. As such, the aforementioned updates only
correspond to approximations in the case where $W$ is non-orthonormal.

%\subsection{Hyper-parameter tuning}
{\color{black}
\paragraph{Hyper-parameter Tuning}

\label{Sec-HypeTune}
In practice, 
the proposed model requires the selection of a single hyper-parameter, $k$, which
determines the dimensionality of  latent variables.
%While
%in some cases it may be possible to select this hyper-parameter based on 
%application specific knowledge, 
We propose to tune $k$ 
by minimizing the negative log-likelihood over 
held-out data. 
%Such an strategy is amenable to the 
%probabilistic nature of the proposed model. 

\section{EXTENSION TO DIRECTED CONNECTIVITY}
%\section{Extension to Bayesian networks}
\label{sec--ExtensionDAG}

In this section we describe a natural extension of the aforementioned model
to estimate causal structure (directed connectivity) across latent variables. 
As in the previous section, we restrict ourselves to  
linear latent variables models where each observed 
variable is conditionally dependent 
on a single latent variable. In the context of Bayesian networks such models
are known as \textit{Pure 1-Factor} models \citep{Silva2006, Kummerfeld2016}. 
We note that such an assumption directly corresponds to each row of the 
loading matrix, $W$, containing at most one non-zero entry. This is precisely what is enforced 
by the non-negativity and orthonormality assumptions introduced in this work.
As such, restricting the loading matrix in this manner corresponds to a natural 
and frequently employed assumption when attempting to recover causal structure 
\citep{Silva2006}. 
%We restrict the loading matrix in this manner in order to recover causal structure, while the 
%aforementioned
%methods study the  higher-order algebraic constraints on covariance structure.
%\textbf{the loading 
%matrix is resticted in this manner in order to recover causal structure, while aforementioned models 
%study higher-order algebraic constraints on covariance structure. }

%\textbf{however, we take a non-Gaussian approach in contrast to the aforementioned models. }

%We assume the following generative model for the 
%We assume that latent variables 
We follow \citet{Shimizu2006} and study a non-Gaussian variant 
of Bayesian networks over latent variables. 
Formally, we assume variables $Z^{(i)}_j, j\in \{1 \ldots k \}$ can be arranged in a causal ordering 
such no later variables causes a variable ahead of it in the order. We denote such an 
ordering by $k(j)$ and assume that each variable, $Z^{(i)}_j$, is a linear function of 
earlier variables together with a non-Gaussian disturbance, $e^{(i)}_j$, such that:
\begin{equation}
\label{SEM_eq}
Z^{(i)}_j = \sum_{k(r) < k(j) }  b^{(i)}_{jr} Z^{(i)}_r + e^{(i)}_j.
\end{equation}
Due to the linear nature of dependencies, we can write equation (\ref{SEM_eq}) as follows:
%\newpage 
%In particular, for the $i$th class we 
%assume:
\begin{align}
\label{DirectStructuralEq}
Z^{(i)} &=   B^{(i)} Z^{(i)} + e^{(i)}\\
%Z^{(i)} 
&=  \left ( I - B^{(i)} \right )^{-1} e^{(i)}.
\end{align}
%where $e^{(i)}$ are non-Gaussian disturbances and $B^{(i)}$ is a 
%matrix which encodes the connection strengths across latent variables. %We assume 
The matrix 
$B^{(i)}$ encodes 
%the \textit{structural model} over latent variables
a Directed Acyclic Graph (DAG), 
which corresponds to the \textit{structural model} over latent variables. 
%such that we may permute 
%its row and columns to make it strictly lower triangular. 
We note that equation (\ref{DirectStructuralEq}) corresponds to a 
Linear, non-Gaussian, acyclic model \citep[LiNGAM;][]{Shimizu2006}. 
%Following \citet{Shimizu2008},
%As before, 
As in the previous section, 
observed data are subsequently related as follows:
\begin{equation}
\label{DirectedMeasurementEq}
X^{(i)} | Z^{(i)} = z^{(i)} \sim \mathcal{N}( W z^{(i)} ,  v^{(i)} I )
\end{equation}
where the loading matrix encodes the \textit{measurement model}. 
To date,
a wide range of algorithms have been proposed to estimate
 measurement models. Prominent examples include the \verb+BuildPureClusters+ 
and \verb+FindOneFactorCluster+ algorithms.  However, such methods  cannot easily be extended to 
the context of multiple related datasets where the underlying 
structural models are heterogeneous. 
%Furthermore, methods such as \verb+FindOneFactorClusters+ do no 
Moreover, such methods do not scale well to high-dimensional data. 
In contrast, our method proposed below can easily accommodate such data and is therefore a good candidate 
to accurately recover the measurement model. 
Once the measurement model has been estimated,  we may proceed 
to infer the causal structure over latent variables using established  methods such as LiNGAM.

%The proposed method is well-suited to the task of recovering the 
%measurement model.  

%\textbf{Add further justification}

We now outline our two-stage procedure to estimate 
\textit{Pure 1-Factor} latent variable models. 
First, the score matching algorithm detailed in Section \ref{Sec--SMalgo}
is employed to estimate the measurement model (i.e., the loading 
matrix $W$). 
%First, the measurement model is estimated by estimating the  
%loading matrix using the score matching algorithm described in 
%Section \ref{Sec--SMalgo}. 
Given a measurement model, there are a variety of algorithms to estimate the
structural moel. In this work we 
follow \citet{Shimizu2008} and 
propose to recover causal dependencies 
over latent variables by applying 
LiNGAM to projected observations, $\hat W^T X^{(i)}$. This step is performed  independently for each of the $N$ classes. 

We note  that the likelihood proposed in Section \ref{sec--LatentConnModel} is misspecified in 
the context of non-Gaussian Bayesian networks considered here (as latent variables follow
non-Gaussian distribution). However, 
in the first stage we are only interested in the estimation of the loading matrix, $W$,
using covariance information which is unaffected by non-Gaussianity over latent variables. 
In fact, as we allow for arbitrary (i.e., non-diagonal) latent connectivities, 
the proposed model is able to accommodate covariances induced by the causal structure
whilst estimating the loading matrix. 
Alternative approaches, such as the \verb+BuildPureClusters+ algorithm, are also 
based exclusively on studying covariance structure, albeit while introducing additional higher-order 
algebraic constraints. % \citep{Silva2006}.

%\textbf{[Consider re-writing..]}
%due to the two-stage 
%nature of our approach, we are only interested in the recovery of the loading matrix, $W$, 
%in the first stage. 

%\newpage 

%\begin{itemize}
%	\item Most methods employ a two-stage approach: first estimate the 
%	measurement model and then estimate the causal model over 
%	latent variables given the measurement model \citep{Silva2006, Kummerfeld2016}
%	
%%	\item Here we focus exclsuive
%	
%	\item Often require assumptions such as \textit{Pure 1-Factor} which indicates that 
%	each observed variable has at most one latent parent. This is exactly the restriction placed on 
%	the loading matrix, $W$, via the non-negativity and orthonormality constraints. 
%	
%	
%\end{itemize}

}

%\newpage 
%\section{Experimental results}
\section{EXPERIMENTAL RESULTS}
\label{sec--Experiments}

We systematically assess the performance of the proposed model using  simulated data
%In the context of simulated data, 
	under the Gaussian factor analysis model of Section \ref{sec--LatentConnModel}
	as well as the 
%We measure the performance when the  data follows a
%Gaussian factor analysis model as well as when the data follows a 
directed latent Pure Bayesian network
%Pure 1-Factor latent variable 
model described in Section \ref{sec--ExtensionDAG}.
Finally, we 
%We also 
present an application to 
resting-state fMRI data  from the ABIDE consortium \citep{DiMartino2014}. 
%In each of the simulations we assess the 
%performance of the proposed method 
%when there is a single ($N=1$) class as well as when there are multiple ($N=10$) classes.
%
%}

%In the context of simulated data, we assess the performance of the proposed 
%method in the context of a single class and multiple related classes. 
%In the context of simulated data, we generate data 
%from the proposed model in order to assess 
%whether we are 
%%if we are 
%able to accurately estimate the loading matrix, $W$, and the 
%covariance structure, $G^{(i)}$, for each subject. 
%In the context of resting-state fMRI data, we 
%assess performance by measuring the negative log-likelihood on held-out data. 

{\color{black}
\subsection{PERFORMANCE METRICS}

We assess the performance of the proposed method 
%In the context of simulated data, we assess the performance of the proposed 
%method 
in the context of both a single class and multiple related classes. 
Throughout these simulations, we quantify the 
performance of various methods based on three distinct tasks:
\begin{enumerate}
	\item Recovery of the loading matrix, $W$. This corresponds to accurately recovering the 
	mixing matrix, and by implication, the module memberships for each variable. 
%	In the context of the 
	\item Recovery of the latent connectivity structure. 
	In the context of undirected latent connectivities, this corresponds to 
	accurately recovering the covariance structure, $G^{(i)}$, across estimated 
	modules.  Conversely, in the context of  directed latent Bayesian network models it 
	corresponds to accurately recovering the causal dependence structure over 
	latent variables, encoded in $B^{(i)}$. 
	
%	, $G^{(i)}$. This corresponds to 
%	accurately recovering the covariance structure across estimated modules. It follows the 
%	good performance in this task is predicated on accurate estimation of the loading matrix,  $W$. 
	\item In the context of 
	undirected connectivities we also 
	measure the negative log-likelihood over unseen data. %Comparing the likelihood for unseen data
	This
	provides an objective quantification of how well the proposed method is able to model the data in comparison 
	to alternative methods. 
	
\end{enumerate}
}
%\newpage 

{\color{black}
%\subsection{Synthetic data}
\subsection{GAUSSIAN FACTOR ANALYSIS MODEL}
\label{sec--GaussFactorSims}
}

\begin{figure*}[t!]
	\centering
	\includegraphics[width=.9\textwidth]{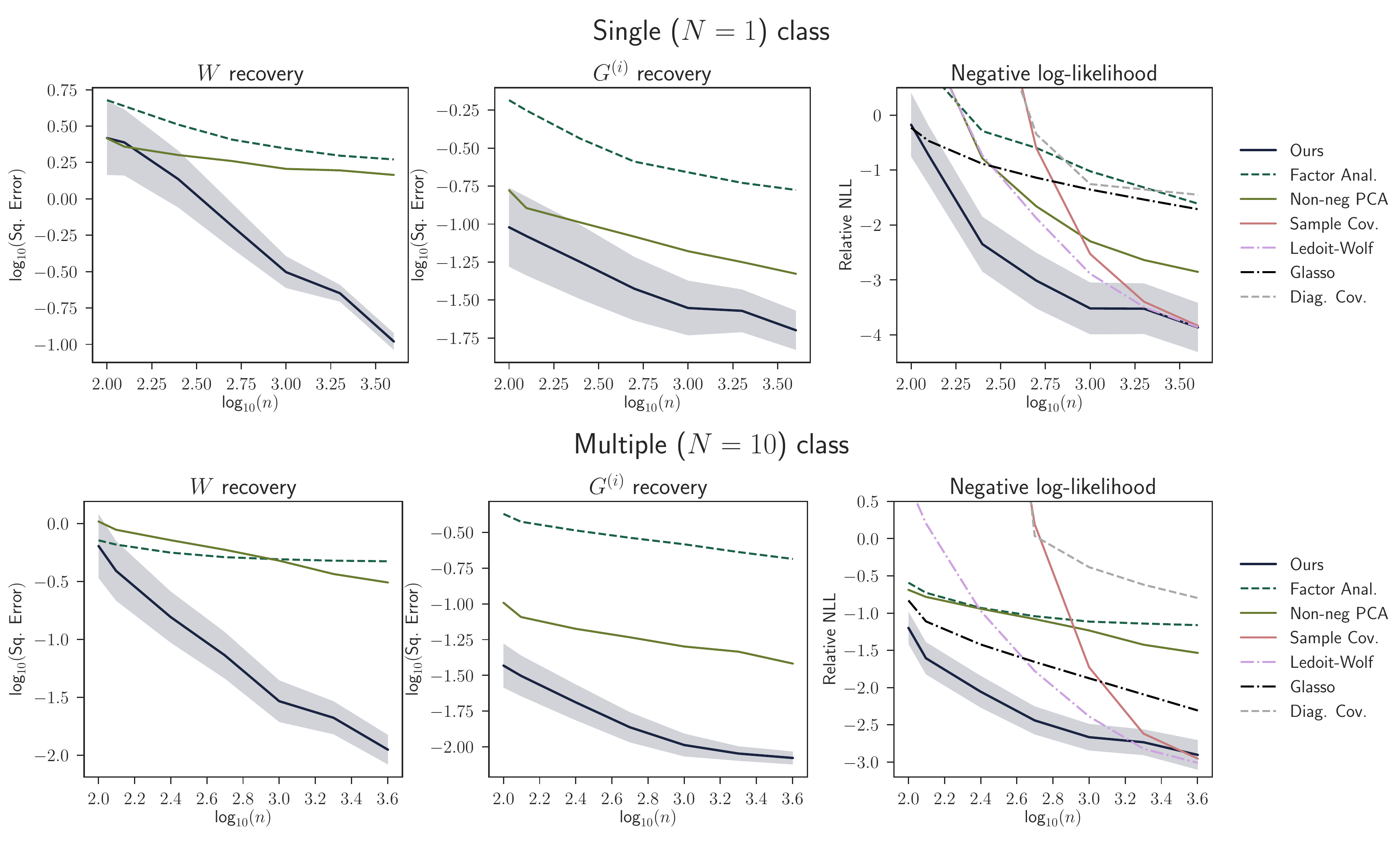}%jointPic_GaussianLatent_extrabase.pdf}%Final_GaussianLatent.pdf}%jointPic_GaussianLatent}%{SingleSubjectCorrect}
	\caption{Simulated data results 
		for Gaussian latent variable models with 
		single ($N=1$) and multiple ($N=10$) classes are
		shown along the top and bottom panels respectively. 
		Left and middle panels plot the mean squared error for the estimated 
		loading and latent variable covariance matrices as a function of sample size, $n$.		
		Right panels shows the mean negative log-likelihood for unseen data as a function of sample size, $n$. 
		Shaded regions correspond to 95$\%$ error bars. 
	}
	\label{fig:SimResultsOneClass}
\end{figure*}
%Data was generated according to the proposed model.

%We perform a simulation study in order to quantify
%how reliably the proposed method is able to recover the 
%mixing matrix, $W$, as well as the covariance structure for latent variables, $G^{(s)}$.
%This was quantified in terms of the squared error between the true and 
%estimated parameters as well as the negative log-likelihood over unseen data. 

%
%
Data was generated according to the  model described in Section \ref{sec--LatentConnModel}. 
The covariance structure for latent variables, $G^{(i)}$, 
%The covariance structure 
was randomly generated for each class by 
%for each class were randomly generated by 
sampling the lower triangular entries from a standard Gaussian distribution and multiplying 
by its transpose. 
We note that generating $G^{(i)}$ in this fashion will randomly introduce 
both positive and negative correlations %marginal dependence 
across modules. %However, in practice, we find that 
%
%Latent variables for each subject were generated according to a zero-mean
%multivariate Gaussian with this covariance structure. 
A random loading matrix, $W$, was generated by sampling uniform random variables and 
projecting onto the non-negative Stiefel manifold (this involved retaining the 
largest entry per row and setting all other entries to zero). 
Observations for each class were subsequently 
generated according to equations 
(\ref{FA_eq1}) and 
(\ref{FA_eq2}). 
The dimensionality of observations and latent variables was set to $p=50$ 
and $k=5$ respectively. 
Data was generated in this manner for $N$ subjects. We consider two cases: $N=1$ and $N=10$, which 
correspond to the single class and multiple class scenarios.  
Each experiment was repeated 500 times. 

The proposed method was benchmarked against several widely used alternatives.
In the context of recovering the loading matrix, $W$, and covariance structure of latent variables, $G^{(i)}$, 
% as well as the covariance 
%structure of latent variables we 
we 
compare to non-negative PCA (using the method 
proposed by \citet{Sigg2008}) and traditional factor analysis (where Varimax 
rotation was employed, since it should be able to recover module structure in the factor loadings).
We note that 
while these methods do not explicitly model the covariance structure across latent variables, this can be 
estimated by first projecting observations using the estimated loading matrix and 
subsequently studying the 
covariance structure. Indeed, this is  related to the update performed by the 
proposed method in equation (\ref{Gupdate}).
When measuring the negative log-likelihood over unseen data we add additional 
comparisons against 
the sample covariance matrix and the %shrinkage 
estimate proposed by  
%Ledoit-Wolf estimate 
\citet{Ledoit} and the graphical Lasso. % \citep{Friedman2008}. 
%and the shrinkage estimator proposed by \cite{Rahim2017}. 
%Furthermore, in the context of multiple related classes we also include the 
%shrinkage estimator proposed by \citet{Rahim2017}. 

%\subsubsection*{Single class case}

Simulation results for a single ($N=1$) class are shown along the 
top panel of 
%Results for the setting of $N=1$ classes 
%are shown in
 Figure \ref{fig:SimResultsOneClass}. 
The top left panel plots the  squared error when estimating the
loading matrix. %, $W$, as a function of the sample size, $n$.
%We note that i
In the presence of small sample sizes, the proposed method is comparable to non-negative 
PCA. However, as the sample size increases the proposed method 
consistently out-performs 
%both non-negative PCA and traditional factor analysis
alternative methods
as it is able to model the connectivity of latent variables. 
Additional results relating the clustering implied by estimated loading matrices
are provided in the supplementary materials. 
Results for the estimation of  latent connectivities, $G^{(i)}$, are shown in the top middle panel. 
We note that the proposed method comfortably outperforms 
competing methods. % for practically all sample sizes.
%
%The proposed method 
%consistently out-performs both non-negative PCA and 
%factor analysis in recovery of the loading matrix and the latent variable covariances.
Finally, 
the right panel shows 
mean negative log-likelihood over unseen data; 
the proposed method 
%In the context of small and moderate sample sizes
%that the proposed method
consistently 
out-performs alternative methods for small and moderate sample sizes and remains 
competitive as sample size increases. 
%Further, 
%additional experiments presented in 
% the supplementary material %we provide further experimental results which 
% demonstrate 
%that the proposed method is able to accuracy cluster variables. 

The bottom panel of 
Figure \ref{fig:SimResultsOneClass} plots results for the general case of multiple subjects. 
While the loading matrix is shared across all classes, the 
latent covariance structure is %highly 
heterogeneous. 
While the proposed method is well-suited to accommodate such data, methods such as 
PCA and factor analysis are not directly applicable. As such, 
non-negative PCA and factor analysis models were applied using
a naive aggregation of the data which 
concatenated observations across all classes. 
%Results for the recovery of the loading matrix are shown in 
%the bottom left panel of Figure \ref{fig:SimResultsOneClass}. 
As before, by 
accurately modeling the marginal dependencies across 
latent variables, the proposed method is able to obtain far more accurate
estimates of both the loading matrix as well as the latent connectivity structure. 
%
%The bottom middle panel shows results for the recovery of the latent connectivities, $G^{(i)}$.
%%As was the case with  previous experiments, 
%The proposed method consistently out-performs 
%traditional approaches by directly modeling the latent connectivity structure. 
%Finally,
The  bottom right panel of Figure \ref{fig:SimResultsOneClass} plots the 
mean negative log-likelihoods over unseen data and provides empirical evidence that 
the proposed model provides an accurate estimate of covariance structure. 
%As before,  proposed 
%method out-performs alternative methods. 

%the mean negative log-likelihood 
%on unseen data. %, with particularly poor performance for non-negative PCA and traditional factor analysis. 

In addition to quantifying the recovery of the loading matrix and latent connectivities, we 
also quantify the computation cost associated with each algorithm. The top panel of Figure 
\ref{fig:RunTime} shows the mean running time as a function of the number of observed variables, $p$. 
The results validate our prior claims that the score matching algorithm yields significant computational
improvements compared to the maximum likelihood algorithm described in the Supplementary material. 
For high-dimensional data, the proposed score matching algorithm also improves on the running 
time compared to both factor analysis and non-negative PCA. 

{\color{black}

%\subsection{Experiments with Pure 1-Factor causal models}
%\subsection{PURE 1-FACTOR BAYESIAN NETWORKS}
\subsection{LATENT PURE BAYESIAN NETWORKS}

\begin{figure*}[th!]
	\centering
	\includegraphics[width=.9\textwidth]{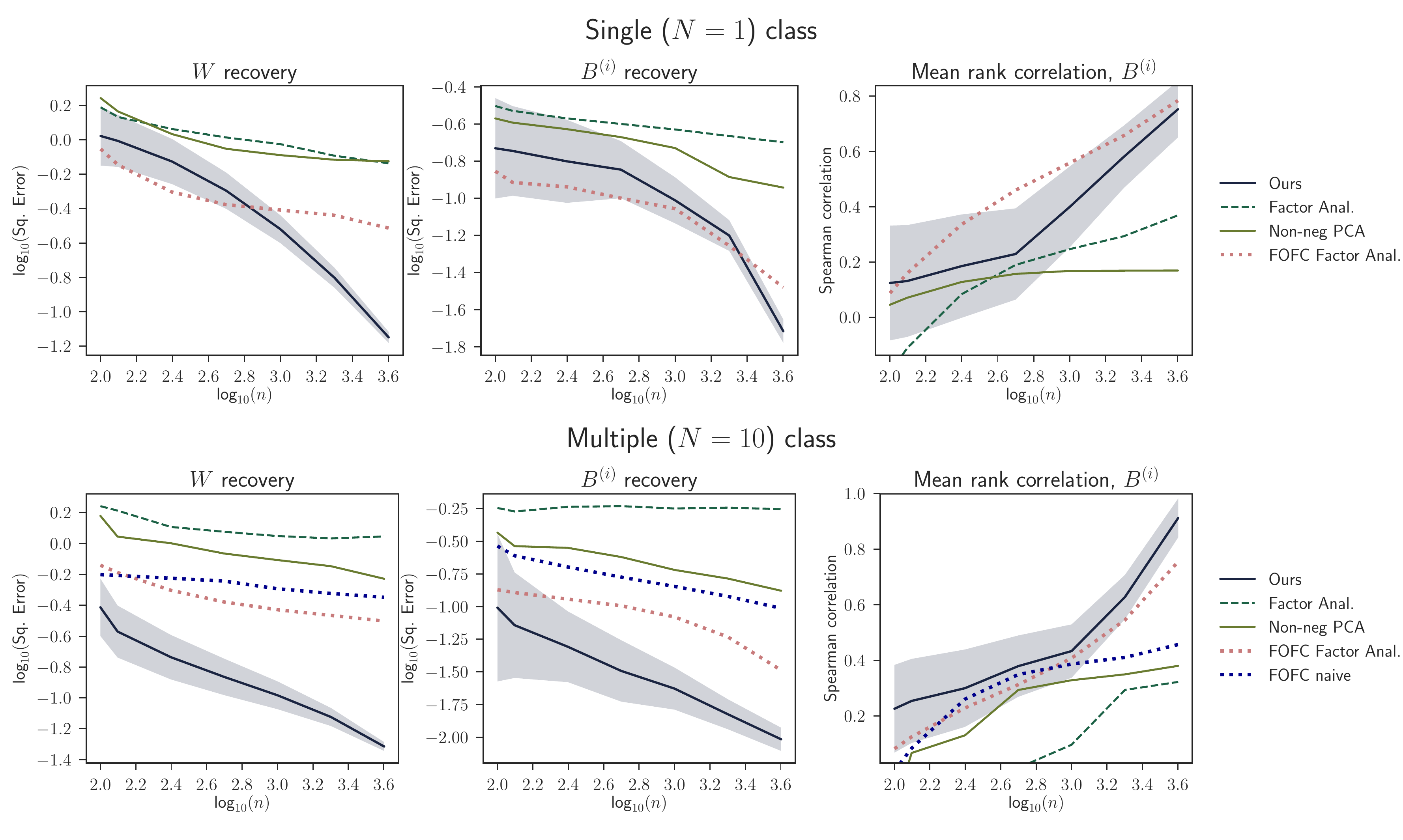}%Final_DirectLatent.pdf}
	\caption{Simulated data results 
		for latent  Bayesian networks with 
		single ($N=1$) and multiple ($N=10$) classes are
		shown along the top and bottom panels respectively. 
		Left and middle panels plot the mean squared error for the estimated 
		loading matrix and structural dependency matrices 
		as a function of sample size, $n$.		
		The right panels show the mean correlation between the estimated causal ordering of latent variables and the true causal order.
		For all algorithms the causal structure over latent variables was estimated by LiNGAM.
		Shaded regions correspond to 95$\%$ error bars. }
	\label{fig:DirectSimRes}
\end{figure*}

In this section 
we perform experiments where the data is generated 
%data was generated
as described in Section \ref{sec--ExtensionDAG}. This involved 
generating latent variables following a non-Gaussian variant of 
Bayesian networks where the disturbances, $e^{(i)}$, were simulated 
according to  a Logistic distribution.  The 
weights for the structural model, encoded in $B^{(i)}$, were randomly
generated together with a distinct random causal ordering  for each class. 
The loading matrix, $W$, was generated as in 
Section \ref{sec--GaussFactorSims}. 

%Throughout these experiments we compared the performance of 
%We benchmark the performance of the 
%two-stage procedure described in Section \ref{sec--ExtensionDAG} with
%alternative methods 
In the context of recovering the measurement model (i.e., the loading matrix $W$)
we benchmark the proposed method with 
factor analysis and non-negative PCA
as well as the 
\verb+FindOneFactorClusters+  (FOFC) algorithm\footnote{The Tetrad project implementation was employed.} proposed by \citet{Kummerfeld2016}. 
Formally, the FOFC algorithm % \verb+FindOneFactorClusters+  
only returns non-overlapping 
clusters of observed variables which share the same latent parent. In order to estimate the 
associated loading matrix we subsequently employ factor analysis whilst preserving 
the 1-Factor structure as suggested by \citet{Shimizu2008}.
%Briefly, the 
%\verb+FindOneFactorClusters+ 
%improves upon the \verb+BuildPureClusters+ algorithm 
Given an  accurate estimate of $W$, 
we may directly project 
observations, $ \hat Z^{(i)} = \hat W^T X^{(i)}$ and apply traditional causal discovery algorithms by
treating $\hat Z^{(i)}$ as observed variables \citep{Silva2006, Shimizu2008}. 
Throughout these experiments, the causal structure of latent variables was inferred 
using LiNGAM.

Figure \ref{fig:DirectSimRes} shows results for a single ($N=1$)
and multiple ($N=10$) class cases along the top and bottom row respectively. 
The left panels show the squared error when estimating the 
loading matrix. 
In the context of causal models this corresponds to accurately recovering the 
measurement model. %, which is shared across all classes.
When data is only available for a single class (top left panel)
the performance of the proposed method is similar to that of the FOFC algoritm.
%the proposed method performs consistently worse that the FOFC % \verb+FindOneFactorClusters+
%algorithm whilst still providing and improvement over factor analysis and non-negative PCA methods. 
However, when data across multiple classes is available, the proposed method 
is able to exploit this information and improve upon the FOFC algorithm as shown in the bottom left panel. 
We also plot the performance of running the FOFC when naively aggregating data 
across multiple subjects. Such a naive aggregation leads to worse performance as 
each class has its own latent causal structure. 

%The middle and right panels show the results 
We observe a similar pattern when studying the recovery of the structural equations, as shown in the middle
and right 
panels.  The proposed method out-performs both factor analysis and PCA and is 
comparable to FOFC in the context of a single ($N=1$) class. However, in the context of multiple 
classes the proposed method is able to out-perform alternative methods.

Finally, the bottom panel of Figure \ref{fig:RunTime} plots the mean running time as the number of 
observed variables, $p$, increases. In terms of running time, the proposed method is 
significantly faster than the FOFC algorithm. 

%The proposed method consistently out-performs 
%factor analysis and PCA (not shown in order to allow for closer comparison of other methods) but is only
%competitive with FOFC based approaches in the context of 
%multiple classes. 
%Finally, the right panels plot the correlation between the true and estimated causal ordering over latent 
%variables measured using the Spearman rank correlation. As the number of observations, $n$, increases
%both the proposed method as well as FOFC algorithms are able to correctly recover the 
%true causal ordering. 

%\newpage 
%\subsection{Application to fMRI data }
\subsection{APPLICATION TO  FMRI DATA}

\begin{figure}[t!]
	\centering
	\includegraphics[width=.45\textwidth]{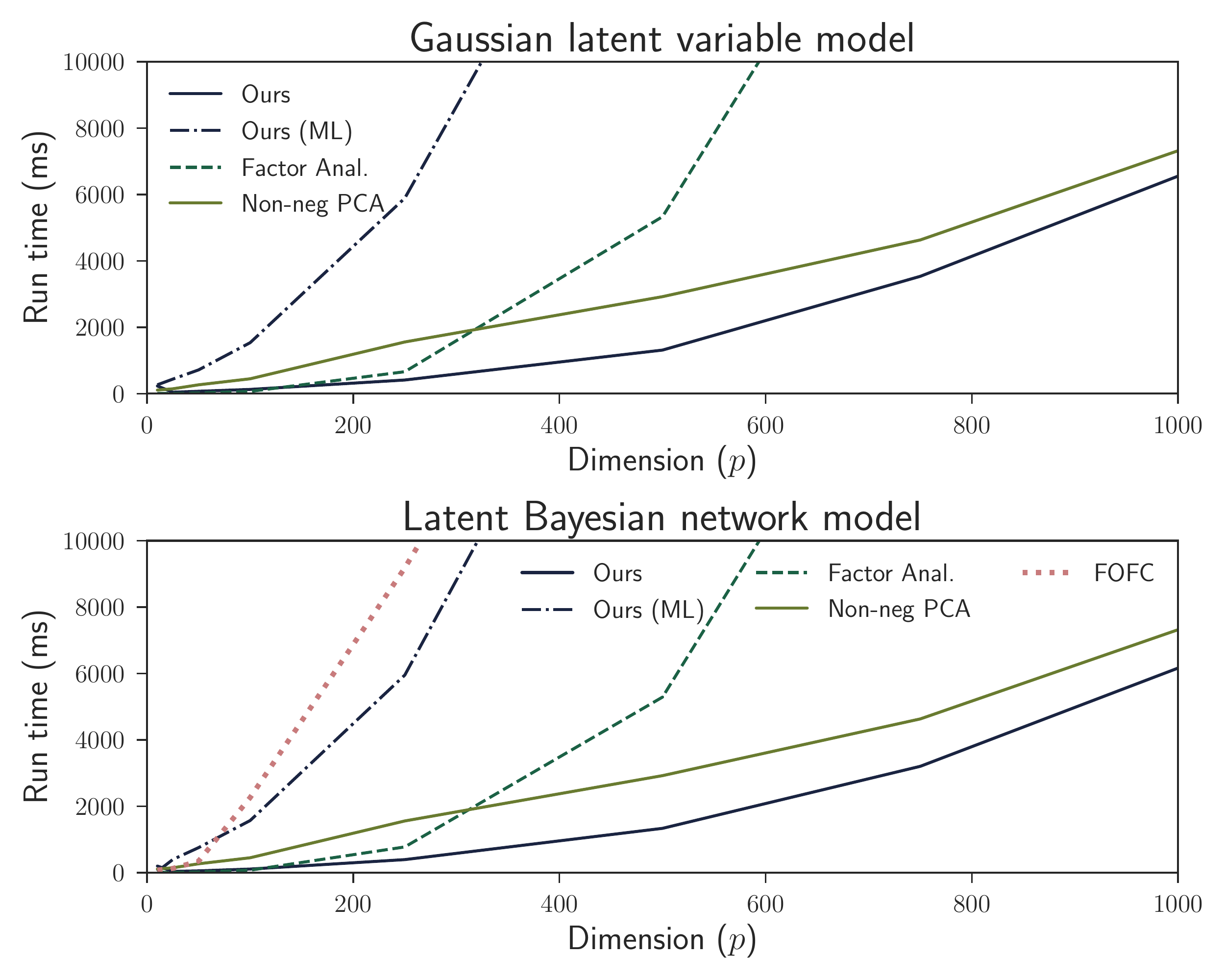}
	\caption{Mean running times (in milliseconds) taken to estimate the 
		factor loading matrix, $W$, when data is generated according to the 
		Gaussian latent variable model (top) and latent Bayesian network model (bottom). 
		Mean run times based on 10 experiments run on a Macbook Pro (3.5 GHz Intel Core i7, 16 GB RAM). 
	}
	\label{fig:RunTime}
\end{figure}

%\newpage 
In this section we 
apply the proposed method to resting-state fMRI data taken from the ABIDE consortium \citep{DiMartino2014}.
Data was collected from the University of Maryland site 
%The ABIDE dataset consists of data
corresponding to 53 healthy controls  as well 
as 
53 age matched Autism Spectrum Disorder (ASD) subjects. %, resulting in a total of 106 subjects.
Data from each subject was treated as a distinct class, resulting in $N=106$ classes.
Data were preprocessed via the CPAC % \url{http://fcp-indi.github.com} for further details } 
pipeline from the ABIDE repository\footnote{http://preprocessed-connectomes-project.org/abide/}. 
%Preprocessing involved slice time correction, motion correction and 
%intensity normalization followed by regression of motion parameters. Linear and quadratic trends were 
%removed from frequency drifts. 
%Mean 
Time courses were then extracted from 116 regions defined by the 
Automated Anatomical Labeling (AAL) atlas, yielding 296 observations over 116 nodes for each subject. 
The  %we analyze the data under the assumption of undirected 
data was analyzed under the assumption of undirected latent connectivity structure
as there is a large literature discussing differences in covariance structure between healthy controls and 
ASD subjects \citep{Fox2010}.

\begin{center}
	{\small
		\begin{table}[b!]
			\small 
			\vskip -.5cm 
			\caption{Mean log-likelihood scores on unseen data for $N=106$ subjects (standard deviations are provided in brackets).}
			\label{Table_LLscore}
			\centering 
			\begin{tabular}{|l l |} 
%				\small 
				\hline
				Method  & Log-likelihood\\ [0.5ex] 
				\hline%\hline
				Ours & 	\textbf{-163.76 (8.86)}\\ 
				Non-neg. PCA &-190.47 (9.26) \\
				Factor Anal.& 	-193.89 (8.05)\\ 
				Glasso &	-198.58 (9.05) \\
				Lediot-Wolf & -247.91 (10.88)\\
				Sample Cov. & -329.75 (14.98)\\
				\hline
			\end{tabular}
		\end{table}
	}
\end{center}

\vskip -.9cm 
The proposed method requires the specification of a single parameter, $k$, which 
dictates the dimensionality of latent variables. 
As discussed in Section \ref{Sec-HypeTune}, 
this parameter was selected by
%we select this parameter by 
minimizing the negative log-likelihood over 
held-out data, resulting in an choice of $k=5$ modules.
%In this section, we divide data into 
%Following \cite{Varoquaux2017a},
%In order to account for the autocorrelated nature of fMRI data as well as 
%inter-subject heterogeneity %\citep{Varoquaux2017a},
%data was  split on a subject-by-subject basis with the 
%% into a training 
%%set consisting of the 
%first 240 observations used for training 
%and the final 56 observations for testing. 
% and a validation set consisting of the final 56 observations.
%This resulted in a 
%The left panel of 
%The data was analyzed under the assumption of undirected latent connectivity structure
%as there is a large literature discussing differences in covariance structure between healthy controls and 
%ASD subjects \citep{Fox2010}.  
%In contrast, differences in causal structure between healthy controls and 
%ASD subjects has not been widely studied. 
%In the context of fMRI research, this allows us to study the connectivity structure 
%in the context of both funcitonal (undirected) and effective (directed) connectivity \citep{Friston2011}.
%
Figure \ref{fig:ABIDEres} shows the $k=5$ estimated 
modules obtained by applying the proposed method. 
%Recall that module structure is encoded by
%the loading matrix, $W$, and is shared 
%across all subjects.
The
spatially consistency and inter-hemispheric symmetry of
module assignments 
reflects the anatomical and functional architecture of the brain. 
Moreover, edges in  Figure \ref{fig:ABIDEres} highlight significant differences in
covariance structure of latent variables between healthy controls and ASD subjects. Permutation tests were performed 
on each edge of the latent connectivity structure and 
%on an edge-by-edge basis and 
Bonferroni corrected for multiple testing. 
%In line with 
Results indicate that ASD subjects demonstrate increased connectivity for which there is 
growing evidence \citep{Keown2013}. 
In particular, we note increased connectivity
between the frontoparietal regions (module 2) and both the
occipital regions (module 1) and the hippocampus, amydala and temporal lobes (module 3).
Finally, Table \ref{Table_LLscore} reports the mean log-likelihood scores on unseen data 
for all $N=106$ subjects, demonstrating the  proposed model
accurately 
captures covariance structure of  fMRI data. 
%The results suggest that modeling fMRI data as 

%The right panel of Figure \ref{fig:ABIDEres} shows the negative log-likelihood over unseen data
%for a variety of different methods. As in the synthetic experiments in Section \ref{sec--GaussFactorSims}, we note that the proposed method out-performs factor analysis and non-negative PCA as well as 
%shrinkage based estimates. 

%%%%%%%%%%%%%%%%%%%%%%%%%%%%%%%%%%%%%%%%%%%%%%%%%%%%%%%%%%%%%%%%%%%%%%%%%%%%
%%%%%%%%%%%%%%%%%%%%%%%%%%%%%%%%%%%%%%%%%%%%%%%%%%%%%%%%%%%%%%%%%%%%%%%%%%%%
%%%%%%%%%%%%%%%%%%%%%%%%%%%%%%%%%%%%%%%%%%%%%%%%%%%%%%%%%%%%%%%%%%%%%%%%%%%%
%%%%%%%%%%%%%%%%%%%%%%%%%%%%%%%%%%%%%%%%%%%%%%%%%%%%%%%%%%%%%%%%%%%%%%%%%%%%

%\newpage 

%\begin{figure*}[t!]
\begin{figure}[t!]
	\centering
	\includegraphics[width=.4\textwidth]{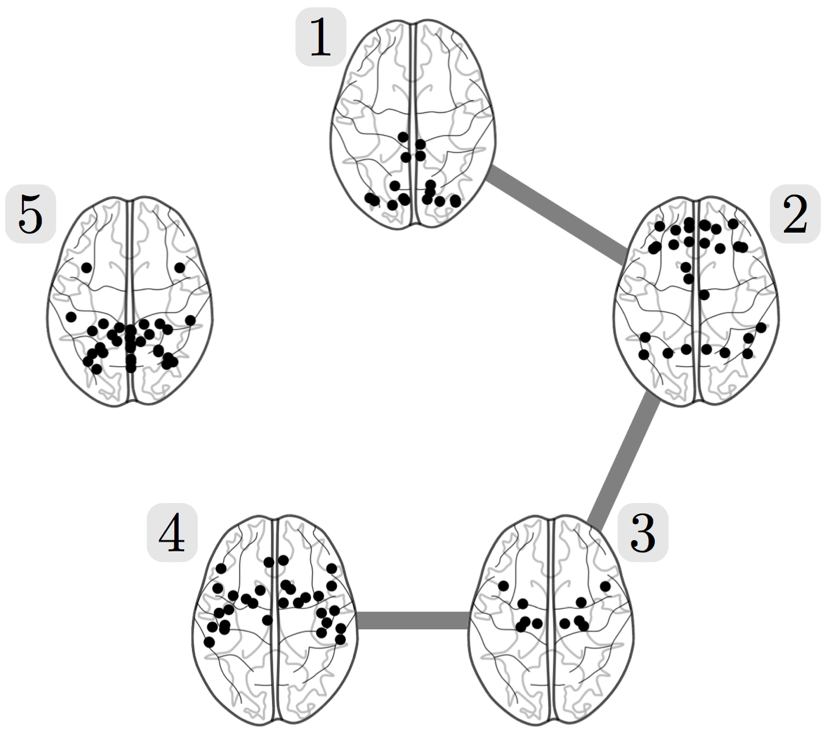}
	\caption{%Left: 
		Visualization of estimation brain modules. 
		%		Estimated modules are visualized; 
		Each black node represents a 
		distinct brain region. 
		%		The regions associated with each of the 5 modules are visualized.
		%		Visualization of the 5 modules estimated by the proposed model. 
		Note that estimated modules are both 
		spatially consistent and symmetric across hemispheres. 
		%		
		%		module assignments 
		%		reflect the anatomical and functional architecture of the brain. 
		%		
		Edges indicate 
		significant increase in inter-module  marginal dependence for ASD subjects compared to 
		healthy controls (edge-wise Bonferroni corrected permutation tests, $p<0.01$). %(based on individual, Bonferroni corrected permutation tests). 
		%		significant differences between inter-module of healthy subjects and ASDestimated usings on individual Bonferroni corrected )
		%		Right: Negative log-likelihood measured over held out (i.e., unseen) data.The proposed method out-performs 
		%		alternative methods.
	}
	\label{fig:ABIDEres}
	%\end{figure*}
\end{figure}

%\section{Conclusion}

%{\scriptsize 
%\begin{center}
%	\begin{tabular}{|c c c c c |} 
%		\hline
%		\textbf{Method}  & Ours & Factor Anal. & Non-neg. PCA & Glasso  \\  
%		\textbf{Log-like.} & -163.76 (8.86) & -193.89 (8.05) & -190.47 (9.26) & -198.58 (9.05)  \\
%		%\hline
%%		Ours & \\ 
%		\hline
%	\end{tabular}
%\end{center}
%}

%Method					Log-likelihood
%Ours					-163.76 (8.86)
%Non-negative PCA		-190.47 (9.26)
%Factor analysis			-193.89 (8.05)
%Graphical Lasso			-198.58 (9.05)
%Ledoit-Wolf				-247.91 (10.88)
%Sample Covariance		-329.75 (14.98)
%Diagonal Covariance		-410.81 (5.66)

\section{CONCLUSION}

We have proposed a probabilistic model which simultaneously performs 
grouping of variables as well as estimation of the
\textit{latent connectivities}  between
groups. The proposed method can be seen as an extension of traditional factor analysis with the 
important difference that 
latent variables are allowed to have a full (i.e., non-diagonal) covariance structure while the loading 
matrix is restricted to encode module membership. 
%The proposed method 
%allows for high-dimensional data to be analyzed in a much more accessible manner and 
%can naturally accommodate data from multiple distinct classes.
%By restricting the loading matrix and allowing for latent variables to have a full covariance structure,
The proposed model can directly accommodate 
%high-dimensional 
datasets across multiple related classes under the assumption that variables across classes share the 
same modularity or community structure.
%In fact, our model is particularly beneficial in context of stu
While the  proposed method is introduced in the context of Gaussian latent variable models, we also demonstrate
that it may be extended to latent Bayesian network models. 
%
%In the context of latent Bayesian networks,
%prior assumptions on the loading matrix can be  understood as enforcing a Pure measurement models.  
%
We present 
experiments on synthetic  and fMRI data  which demonstrate the capabilities of our approach, in particular
showing it successfully scales to high-dimensional data.}
%, both in the 
%context of directed and undirected latent variables. 
%We further present 
%An application to resting-state fMRI data 
% identifies plausible differences in connectivity structure between healthy controls and ASD subjects. 

\newpage 

{\normalsize
\bibliography{library_new}

\begin{thebibliography}{31}
\providecommand{\natexlab}[1]{#1}
\providecommand{\url}[1]{\texttt{#1}}
\expandafter\ifx\csname urlstyle\endcsname\relax
  \providecommand{\doi}[1]{doi: #1}\else
  \providecommand{\doi}{doi: \begingroup \urlstyle{rm}\Url}\fi

\bibitem[Airoldi et~al.(2008)]{Airoldi2008}
Airoldi, Edoardo et~al.
\newblock {Mixed Membership Stochastic Blockmodels}.
\newblock \emph{J. Mach. Learn. Res.}, 9\penalty0 (2008):\penalty0 1981--2014,
  2008.
\newblock ISSN 1532-4435.

\bibitem[Bertsekas(2014)]{Bertsekas2014}
Bertsekas, Dimitri~P.
\newblock \emph{{Constrained optimization and Lagrange multiplier methods}}.
\newblock Academic Press, 2014.

\bibitem[Buesing et~al.(2014)]{Buesing2014}
Buesing, Lars et~al.
\newblock {Clustered factor analysis of multineuronal spike data}.
\newblock \emph{NIPS}, 27:\penalty0 3500--3508, 2014.

\bibitem[Comon(1994)]{Comon1994}
Comon, Pierre.
\newblock {Independent component analysis - a new concept? Signal Processing}.
\newblock \emph{Signal Processing}, 36:\penalty0 287--314, 1994.

\bibitem[Damoiseaux et~al.(2006)]{Damoiseaux2006}
Damoiseaux, J~S et~al.
\newblock {Consistent resting-state networks across healthy subjects}.
\newblock \emph{Proc. Natl. Acad. Sci.}, 103\penalty0 (37):\penalty0
  13848--13853, 2006.
\newblock ISSN 0027-8424.

\bibitem[Danaher et~al.(2014)]{Danaher2014}
Danaher, Patrick et~al.
\newblock {The joint graphical lasso for inverse covariance estimation across
  multiple classes}.
\newblock \emph{J. R. Stat. Soc. Ser. B Stat. Methodol.}, 76\penalty0
  (2):\penalty0 373--397, 2014.

\bibitem[Dempster(1972)]{Dempster1972}
Dempster, A~P.
\newblock {Covariance Selection}.
\newblock \emph{Biometrics}, 28\penalty0 (1):\penalty0 157, 1972.

\bibitem[{Di Martino} et~al.(2014)]{DiMartino2014}
{Di Martino}, Adriana et~al.
\newblock {The autism brain imaging data exchange: Towards a large-scale
  evaluation of the intrinsic brain architecture in autism}.
\newblock \emph{Mol. Psychiatry}, 19\penalty0 (6):\penalty0 659--667, 2014.
\newblock ISSN 14765578.

\bibitem[Donoho \& Stodden(2004)Donoho and Stodden]{Donoho2004}
Donoho, David and Stodden, Victoria.
\newblock {When does non-negative matrix factorization give a correct
  decomposition into parts?}
\newblock \emph{NIPS}, pp.\  1141--1148, 2004.

\bibitem[Fox \& Greicius(2010)Fox and Greicius]{Fox2010}
Fox, Michael~D. and Greicius, Michael.
\newblock {Clinical applications of resting state functional connectivity}.
\newblock \emph{Front. Syst. Neurosci.}, 4, 2010.

\bibitem[Hirayama et~al.(2016)]{Hirayama2016}
Hirayama, Jun~Ichiro et~al.
\newblock {Characterizing variability of modular brain connectivity with
  constrained principal component analysis}.
\newblock \emph{PLoS One}, 11\penalty0 (12), 2016.

\bibitem[Hyv{\"{a}}rinen(2005)]{Hyvarinen2006}
Hyv{\"{a}}rinen, Aapo.
\newblock {Estimation of non-normalized statistical models by score matching}.
\newblock \emph{J. Mach. Learn. Res.}, 6:\penalty0 695--708, 2005.

\bibitem[Hyv{\"{a}}rinen(2007)]{Hyvarinen2007}
Hyv{\"{a}}rinen, Aapo.
\newblock {Some extensions of score matching}.
\newblock \emph{Comput. Stat. Data Anal.}, 51\penalty0 (5):\penalty0
  2499--2512, 2007.

\bibitem[Keown et~al.(2013)]{Keown2013}
Keown, Christopher~Lee et~al.
\newblock {Local functional overconnectivity in posterior brain regions is
  associated with symptom severity in autism spectrum disorders}.
\newblock \emph{Cell Rep.}, 5\penalty0 (3):\penalty0 567--572, 2013.

\bibitem[Kummerfeld \& Ramsey(2016)Kummerfeld and Ramsey]{Kummerfeld2016}
Kummerfeld, Erich and Ramsey, Joseph.
\newblock {Causal Clustering for 1-Factor Measurement Models}.
\newblock \emph{KDD}, pp.\  1655--1664, 2016.

\bibitem[Ledoit \& Wolf(2003)Ledoit and Wolf]{Ledoit}
Ledoit, Olivier and Wolf, Michael.
\newblock {Improved estimation of the covariance matrix of stock returns with
  an application to portfolio selection}.
\newblock \emph{J. Empir. Financ.}, 10\penalty0 (5):\penalty0 603--621, 2003.

\bibitem[Lin et~al.(2016)]{Lin2016}
Lin, Lina et~al.
\newblock {Estimation of high-dimensional graphical models using regularized
  score matching}.
\newblock \emph{Electron. J. Stat.}, 10:\penalty0 806--854, 2016.

\bibitem[Marlin \& Murphy(2009)Marlin and Murphy]{Marlin2009}
Marlin, Benjamin and Murphy, Kevin.
\newblock {Sparse Gaussian Graphical Models with Unknown Block Structure}.
\newblock \emph{ICML}, pp.\  705--712, 2009.

\bibitem[Monti et~al.(2017)]{Monti2017}
Monti, Ricardo~Pio et~al.
\newblock {Learning population and subject-specific brain connectivity networks
  via Mixed Neighborhood Selection}.
\newblock \emph{Ann. Appl. Stat.}, 11\penalty0 (4):\penalty0 2142--2164, 2017.

\bibitem[Newman(2006)]{Newman2006}
Newman, M E~J.
\newblock {Modularity and community structure in networks}.
\newblock \emph{Proc. Natl. Acad. Sci.}, 103\penalty0 (23):\penalty0
  8577--8582, 2006.

\bibitem[Paatero \& Tapper(1994)Paatero and Tapper]{Paatero1994}
Paatero, Pentti and Tapper, Unto.
\newblock {Positive matrix factorization: A non-negative factor model with
  optimal utilization of error estimates of data values}.
\newblock \emph{Environmetrics}, 5\penalty0 (2):\penalty0 111--126, 1994.

\bibitem[Pearl(2009)]{Pearl2009}
Pearl, Judea.
\newblock \emph{{Causality}}.
\newblock Cambridge University Press, 2009.

\bibitem[Sasaki et~al.(2017)]{Sasaki17NC}
Sasaki, Hiroaki et~al.
\newblock Simultaneous estimation of non-gaussian components and their
  correlation structure.
\newblock \emph{Neural Comput.}, 29:\penalty0 2887--2924, 2017.

\bibitem[Seung \& Lee(1999)Seung and Lee]{Seung1999}
Seung, H.~Sebastian and Lee, Daniel~D.
\newblock {Learning the parts of objects by non-negative matrix factorization}.
\newblock \emph{Nature}, 401\penalty0 (6755):\penalty0 788--791, 1999.

\bibitem[Shimizu et~al.(2006)]{Shimizu2006}
Shimizu, Shohei et~al.
\newblock {A Linear Non-Gaussian Acyclic Model for Causal Discovery}.
\newblock \emph{J. Mach. Learn. Res.}, 7:\penalty0 2003--2030, 2006.

\bibitem[Shimizu et~al.(2009)]{Shimizu2008}
Shimizu, Shohei et~al.
\newblock {Estimation of linear non-Gaussian acyclic models for latent
  factors}.
\newblock \emph{Neurocomputing}, 72\penalty0 (7-9):\penalty0 2024--2027, 2009.

\bibitem[Sigg \& Buhmann(2008)Sigg and Buhmann]{Sigg2008}
Sigg, Christian~D and Buhmann, Joachim~M.
\newblock {Expectation-maximization for sparse and non-negative PCA}.
\newblock \emph{ICML}, pp.\  960--967, 2008.

\bibitem[Silva et~al.(2006)]{Silva2006}
Silva, Ricardo et~al.
\newblock {Learning the structure of linear latent variable models}.
\newblock \emph{J. Mach. Learn. Res.}, 7:\penalty0 191--246, 2006.

\bibitem[Sporns \& Betzel(2016)Sporns and Betzel]{Sporns2016}
Sporns, Olaf and Betzel, Richard~F.
\newblock {Modular Brain Networks}.
\newblock \emph{Annu. Rev. Psychol.}, 67\penalty0 (1):\penalty0 613--640, 2016.

\bibitem[Tipping \& Bishop(1999)Tipping and Bishop]{Tipping1999}
Tipping, M.~E. and Bishop, C.~M.
\newblock {Probablistic Principle Component Analysis}.
\newblock \emph{J. R. Stat. Soc. Ser. B (Statistical Methodol.}, 11\penalty0
  (2):\penalty0 150--210, 1999.

\bibitem[Varoquaux et~al.(2010)]{Varoquaux2010}
Varoquaux, Ga{\"{e}}l et~al.
\newblock {Brain covariance selection: better individual functional
  connectivity models using population prior}.
\newblock \emph{NIPS}, 2010.

\end{thebibliography}
\bibliographystyle{icml2018}
}
\onecolumn
\section*{Supplementary Material}

\subsection*{Maximum likelihood estimation}

%\subsection*{Maximium likelihood estimation of the proposed model}

In this supplement we derive the 
maximum likelihood estimation algorithm for the proposed 
Gaussian factor analysis model. 
We note that the log-likelihood associated with the proposed model is:
\begin{eqnarray}
\mathcal{L} = \sum_{i=1}^N ~p \log 2 \pi + \log \mbox{det} ~ \Sigma^{(i)}  + \mbox{tr} \left (  {\Sigma^{(i)}}^{-1} K^{(i)} \right ).
\end{eqnarray}
In the case of the loading matrix, the gradient update is defined as:
\begin{align}
\label{ML_wupdate}
\frac{\partial \mathcal{L}}{\partial W } &= \sum_{i=1}^N \frac{\partial \mathcal{L}}{\partial \Sigma^{(i)}}   \frac{\partial \Sigma^{(i)} }{\partial W } = \sum_{i=1}^N \left ( \underbrace{ - {\Sigma^{(i)}}^{-1} + {\Sigma^{(i)}}^{-1} K^{(i)} {\Sigma^{(i)}}^{-1} }_{M^{(i)}}\right ) W G^{(i)} % \\
%&= \sum_{i=1}^N \left ( -{v^{(i)}}^{-1}  I + {v^{(i)}}^{-1}  W A^{(i)} W^T + {v^{(i)}}^{-2} K^{(i)} 
%-2 {v^{(i)}}^{-2} W A^{(i)} W^T K^{(i)}    \right )
\end{align}
We note that the main computational burden is associated with computing $M^{(i)}$. 
Using the Sherman-Woodbury identity, we may write $M^{(i)}$ as:
\begin{align*}
M^{(i)} =&  -{v^{(i)}}^{-1}  I + {v^{(i)}}^{-1}  W A^{(i)} W^T + {v^{(i)}}^{-2} K^{(i)} \\
&-2 {v^{(i)}}^{-2} W A^{(i)} W^T K^{(i)}    \\
&+ {v^{(i)}}^{-2} W A^{(i)} W^T K^{(i)} W A^{(i)} W^T ,
\end{align*}
from which it 
%It thereby 
follows that computing the gradient of the log-likelihood with respect to the loading matrix, $W$, incurs a computational cost of $\mathcal{O}(p^3)$. 

%We note that computing equation(\ref{ML_wupdate}) incurs a computational cost of $\mathcal{O}(p^3)$. 

In the case of the latent connectivity matrix, $G^{(i)}$, the update is equivalent to the score matching algorithm. 
This follows from:
\begin{align}
\frac{\partial \mathcal{L}}{\partial G^{(i)} } &= \sum_{i=1}^N \frac{\partial \mathcal{L}}{\partial \Sigma^{(i)}}   \frac{\partial \Sigma^{(i)} }{\partial G^{(i)} } \\
&=  \sum_{i=1}^N \left ( - {\Sigma^{(i)}}^{-1} + {\Sigma^{(i)}}^{-1} K^{(i)} {\Sigma^{(i)}}^{-1} \right ) W^TW.
\label{G_mlupdate}
\end{align}
Setting equation (\ref{G_mlupdate}) to equal zero implies that $I = K^{(i)} {\Sigma^{(i)}}^{-1} $, which 
after re-arranging yields:
\begin{equation}
G^{(i)} = W^T K^{(i)} W - v^{(i)} I . 
\label{ML_G_update}
\end{equation}

\subsection*{Score matching estimation}

In this supplement we provide a detailed derivation for the score matching algorithm 
presented in Section 4. 
We begin by explicitly writing the score matching objective in terms of parameters $W$, $\{G^{(i)}\}$ and $\{v^{(i)}\}$. 
This is specified as:
\begin{align}
J &= \sum_{i=1}^N - \mbox{tr} \left ( \Omega^{(i)} \right ) + \frac{1}{2} \mbox{tr} \left ( \Omega^{(i)}  \Omega^{(s)}  K^{(i)} \right ) \\
&= \sum_{i=1}^N \left [ - {v^{(i)}}^{-1}\mbox{tr} \left (   I \right ) + {v^{(i)}}^{-1} \mbox{tr}( A^{(i)})   + \frac{1}{2} {v^{(i)}}^{-2} \mbox{tr} (K^{(i)}) \right . \\
&~~~~~~~~~~\left . - {v^{(i)}}^{-2} \mbox{tr} (W^T K^{(i)} W A^{(i)} ) + \frac{1}{2} {v^{(i)}}^{-2} \mbox{tr} (W^T K^{(i)} W A^{(i)} A^{(i)}) \right ]
\end{align}
where $A^{(i)} = G^{(i)} ( G^{(i)}  + v^{(i)}I)^{-1} $ as in the original text. 
We may now directly compute the derivatives with respect to each of the parameters in the proposed latent variable model.
The derivative for the loading matrix is:
\begin{align}
\frac{\partial J}{\partial W} = \sum_{i=1}^N  {v^{(i)}}^{-2} K^{(i)} W \left ( \frac{1}{2} A^{(i)} A^{(i)} - A^{(i)} \right ).
\end{align}

The derivative with respect to latent connectivities can be obtained via the chain rule as:
\begin{align}
\frac{\partial J}{\partial G^{(i)}} &= \frac{\partial J}{\partial A^{(i)}}  \frac{\partial A^{(i)}}{\partial G^{(i)}} \\
&= {v^{(i)}}^{-2} \left [  {v^{(i)}} I -   W^T K^{(i)} W\left ( I - A^{(i)} \right )  \right ] \frac{\partial A^{(i)}}{\partial G^{(i)}} \\
&= {v^{(i)}}^{-2} \left [  {v^{(i)}} I -   W^T K^{(i)} W \left ( I - A^{(i)}\right )  \right ] \left [   ( G^{(i)} + v^{(i)} I  )^{-1} \left ( I - A^{(i)}\right ) \right ].
\label{FinalEqGupdateSM}
\end{align}
We note that setting equation (\ref{FinalEqGupdateSM}) to zero implies that the middle term must be zero, as both 
$ ( G^{(i)} + v^{(i)} I  )^{-1}$  and $(I - A^{(i)})$ cannot be zero. % as $G^{(i)}$ is positive, semi-definite and 
By equating the middle term with zero, we obtain:
\begin{align}
v^{(i)} \left ( W^T K^{(i)} W \right )^{-1} = I - A^{(i)}, 
\end{align}
which after re-arranging yields:
\begin{align}
A^{(i)} &= G^{(i)} ( G^{(i)} + v^{(i) } I)^{-1} = I  -  v^{(i)} \left ( W^T K^{(i)} W \right )^{-1} .
%G^{(i)} ( G^{(i)} + v^{(i) } I)^{-1} &= I  -  v^{(i)} \left ( W^T K^{(i)} W \right )^{-1} 
\end{align}
Finally, we may re-arranging for $G^{(i)}$ to obtain: 
\begin{equation}
G^{(i)} = W^T K^{(i)} W - v^{(i)}  I
\end{equation}
which is the same update as obtained in equation (\ref{ML_G_update}) above.
Before deriving the derivative of the score matching objective with respect to $v^{(i)}$, we state the following identities which will of 
use later on:
\begin{itemize}
	\item We may eigendecompose the latent connectivity $G^{(i)}$ as follows:
	\begin{equation}
	G^{(i)} = V_i D_i V_i^T,
	\end{equation}
	where $V_i$ is a matrix of eigenvectors and $D_i$ is a diagonal matrix of eigenvalues $d_1, \ldots, d_k$. 
	Therefore we may write $A^{(i)}$ as follows:
	\begin{align*}
	A^{(i)} &= G^{(i)} ( G^{(i)} + v^{(i)} I )^{-1}\\
	&= V_i ~D_i~  V_i^T ~V_i ~ \mbox{diag} \left ( \frac{1}{d_j + v^{(i)}}  \right ) ~  V_i ^T \\
	&= V_i  ~ \mbox{diag} \left ( \frac{d_j}{d_j + v^{(i)}}  \right ) ~  V_i ^T
	\end{align*}
	As a result, we can compute the derivative of $A^{(i)}$ with respect to $v^{(i)}$ as follows:
	\begin{equation}
	\label{D_tilde}
	\frac{ \partial A^{(i)}}{\partial v^{(i)} } = V_i ~ \mbox{diag} \left ( \frac{-d_j}{ (d_j + v^{(i)})^2  }   \right ) ~ V_i^T = \tilde D_i .
	\end{equation}
	Furthermore, because $V_i$ are eigenvectors, we have that $	\frac{ \partial \mbox{tr}(A^{(i)})}{\partial v^{(i)} } = \mbox{tr}(\tilde{  {D_i } })$.
	\item 	Using the same arguments, we may write the derivative of $A^{(i)} A^{(i)}$ with respect to $v^{(i)}$ as follows:
	\begin{equation}
	\label{D_double_tilde}
	\frac{ \partial A^{(i)} A^{(i)}}{\partial v^{(i)} } = V_i^T ~ \mbox{diag} \left ( \frac{-2d_j^2 }{ (d_j + v^{(i)})^3  }   \right ) ~ V_i^T = \tilde{ \tilde {D_i } }
	\end{equation}
\end{itemize}

\noindent
Using equations (\ref{D_tilde}) and (\ref{D_double_tilde}) we may therefore write the derivative of the score matching 
objective with respect to $v^{(i)}$ as follows:

%\begin{align}
%\frac{\partial J }{\partial v^{(i)}} = &~~ {v^{(i)}}^{-3}  \mbox{tr} \left (  v^{(i)} I - K^{(i)}\right ) \\
%&+ {v^{(i)}}^{-2}  \mbox{tr} \left ( W^T K^{(i)} W \left [  2 A^{(i)}  - \tilde{D_i}  - {v^{(i)}}^{-1} A^{(i)}A^{(i)} + \frac{1}{2} \tilde{\tilde{D_i}}  \right ] \right  ) \\
%&- {v^{(i)}}^{-1} \mbox{tr} \left( A^{(i)} - v^{(i)} \tilde{D_i} \right)
%\end{align}

\begin{align}
\frac{\partial J }{\partial v^{(i)}} = &~~ {v^{(i)}}^{-3}  \mbox{tr} \left (  v^{(i)} I - K^{(i)}\right ) \\
&+ {v^{(i)}}^{-3}  \mbox{tr} \left ( W^T K^{(i)} W \left [  2 A^{(i)}  - v^{(i)} \tilde{D_i}  -  A^{(i)}A^{(i)} + v^{(i)} \tilde{\tilde{D_i}}  \right ] \right  ) \\
&- {v^{(i)}}^{-2} \mbox{tr} \left( A^{(i)} - v^{(i)} \tilde{D_i} \right)
\end{align}

\noindent 
As such, we define:
\begin{align*}
H^{(i)}_1 &= 2 A^{(i)}  - v^{(i)} \tilde{D_i}  -  A^{(i)}A^{(i)} + v^{(i)} \tilde{\tilde{D_i}} \\
H^{(i)}_2 &= - {v^{(i)}}^{-2} \mbox{tr} \left( A^{(i)} - v^{(i)} \tilde{D_i} \right )
\end{align*}

%\newpage 

\subsection*{Additional experiments: cluster recovery}
\begin{figure}[h]
	\centering
	\includegraphics[width=.6\textwidth]{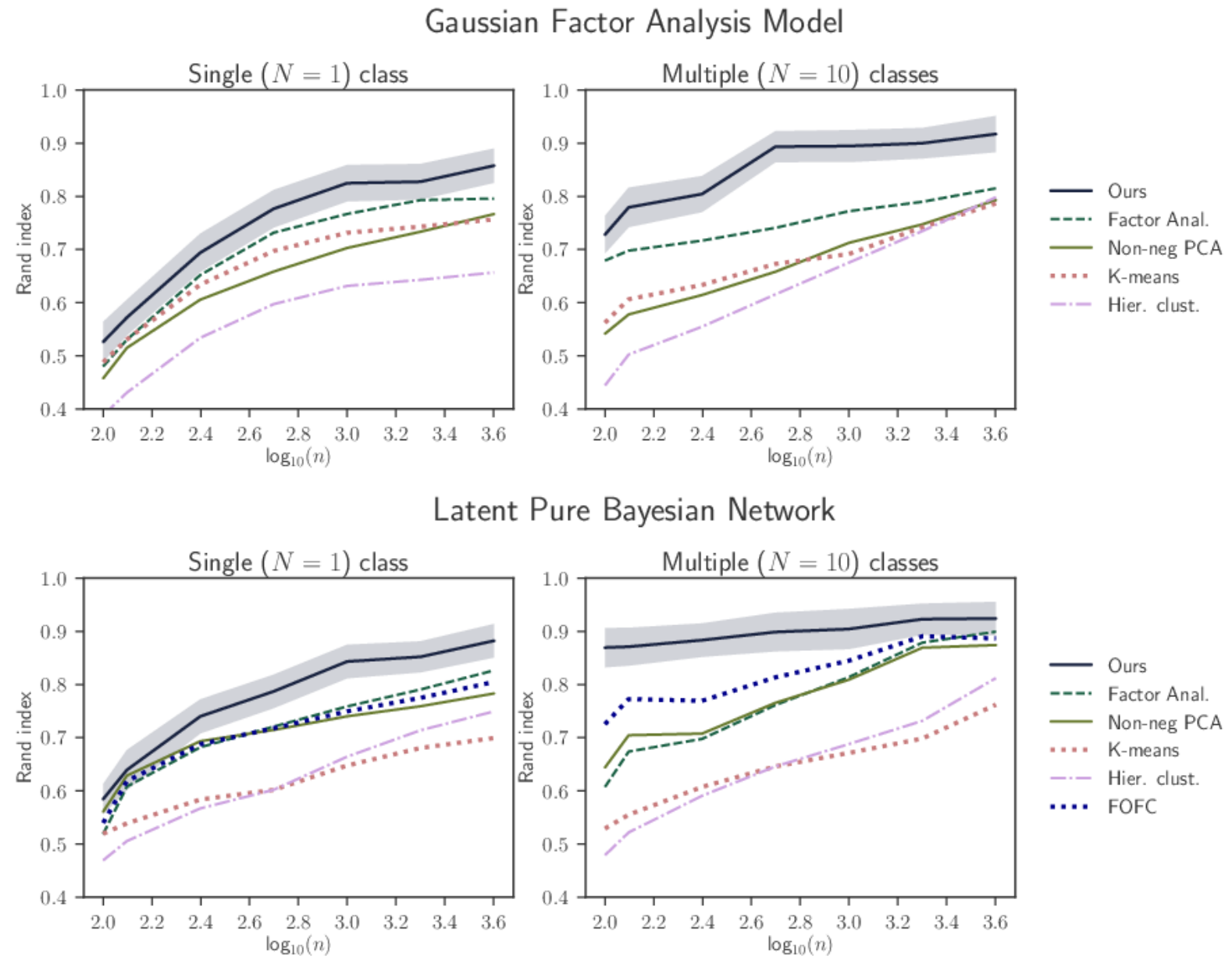}
	\caption{Adjusted Rand index scores for 
		variable clustering as inferred by the estimated loading matrices. Results are shown for 
		Gaussian factor analysis models (top) and latent Bayesian networks (bottom) as well as for 
		$N=1$ and $N=10$ classes in the left and right columns respectively. 
		Shaded regions correspond to $95\%$ error bars. 
	}
	\label{fig:Clustering}
\end{figure}

Results in Section 6 demonstrated that the proposed method is able to reliably 
recover the loading matrix, $W$, in terms of mean squared error.
%However, we are
%also interested to see if the 
In this section we provide additional  results demonstrating that the 
clusters inferred from the estimated loading matrix accurately reflect the true clustering of 
variables. 
The adjusted Rand Index was employed in order to quantify the 
similarity between the estimated and true clusterings. Results are shown in Figure \ref{fig:Clustering}
for Gaussian factor analysis models and latent Bayesian networks along the top and bottom 
row respectively. 
In each case, we note that the proposed model is able to accurately cluster variables, inclusively when
compared with traditional clustering algorithms such as $k$-means and hierarchical clustering. 
Furthermore, we note that as the number of classes increases from $N=1$ to $N=10$, the
accuracy of the proposed method improves as it is able to combine observations across classes. 

\label{additonalClusteringExp}

\end{document}